\documentclass{article}
\usepackage{spconf,amsmath,graphicx}
\usepackage{multirow}
\usepackage{cite}
\usepackage{xspace}
\usepackage{bm}
\usepackage{siunitx}
\usepackage{booktabs}
\usepackage{tabularx}

\ninept

\def\F0{$F_0$\xspace}
\def\textbased{\textbf{text-based}\xspace}
\def\fzbased{$\bm{F_0}$-\textbf{based}\xspace}
\def\shuffle{\textbf{shuffle}\xspace}
\def\daft{\textbf{Daft-Exprt}\xspace}

\title{Do Prosody Transfer Models Transfer Prosody?}

\name{Atli Thor Sigurgeirsson\thanks{This work was supported in part by Huawei and the UKRI Centre for Doctoral Training in Natural Language Processing, funded by the UKRI (grant EP/S022481/1) and the University of Edinburgh, School of Informatics and School of Philosophy, Psychology \& Language Sciences.
}, Simon King}
\address{The Centre for Speech Technology Research, University of Edinburgh, UK}

\begin{document}

\maketitle

\begin{abstract}
 Some recent models for Text-to-Speech synthesis aim to transfer the prosody of a reference utterance to the generated target synthetic speech. This is done by using a learned embedding of the reference utterance, which is used to condition speech generation. During training, the reference utterance is identical to the target utterance. Yet, during synthesis, these models are often used to transfer prosody from a reference that differs from the text or speaker being synthesized.
 
 To address this inconsistency, we propose to use a different, but prosodically-related, utterance during training too. We believe this should encourage the model to learn to transfer only those characteristics that the reference and target have in common. If prosody transfer methods do indeed transfer prosody they should be able to be trained in the way we propose. However, results show that a model trained under these conditions performs significantly worse than one trained using the target utterance as a reference. To explain this, we hypothesize that prosody transfer models do not learn a transferable representation of prosody, but rather a utterance-level representation which is highly dependent on both the reference speaker and reference text.
\end{abstract}


\section{Introduction}
Neural text-to-speech (TTS) has reached a high degree of perceived naturalness by leveraging large neural networks and vast amounts of speech data \cite{sotelo2017char2wav, gibiansky2017deep, shen2018natural, ren2020fastspeech}. Most neural end-to-end models implicitly model prosody. That is, the model learns to infer prosodic features directly from text without any explicit supervision. Prosody is a prominent factor in conveying both emotion and meaning and is therefore necessary for generating natural, expressive speech \cite{wagner2010experimental}. But prosody synthesized by most end-to-end TTS models is an \textit{average prosody}, reflecting the broad statistics of the training data. The mapping from text to prosody is one-to-many: the same text can be spoken in multiple perfectly reasonable prosodic realizations \cite{wilson2006relevance}. By augmenting the verbal component with acoustic correlates of prosody \cite[p.\ 13]{cole2015prosody}, speakers can vary meaning, attitude toward the subject or the listener. End-to-end TTS models on the other hand do not generally select amongst possible prosodic renditions: each input text produces just one prosodic rendition. 
Explicit prosody modeling addresses both the low prosodic variance and the lack of prosody control in end-to-end TTS. One technique is \textit{prosody transfer} (PT) which was first described in \cite{pt_main} and has subsequently been widely studied \cite{pt_main,akuzawa2018expressive,battenberg2019effective, Lee_Kim_2019,zhang2019learning,klimkov2019fine, karlapati2020copycat,zaidi2021daft}. PT models are trained to \textit{transfer} prosody from a reference to a target utterance. During training, PT models are conditioned on the target acoustics (usually the mel-spectrogram) to influence target prosody. During inference, any reference can be used: it can be spoken by the target-speaker or some other speaker (\textit{different-speaker PT}), and the reference can contain the same verbal information or not (\textit{different-text PT}). PT models are therefore tasked with capturing a transferable representation of the reference while preserving the target verbal component and target-speaker identity.

PT models have been shown to capture prosodic information and to have the capacity to produce prosodic variance beyond that found in the training data \cite{pt_main, zaidi2021daft, karlapati2020copycat}. They are therefore capable of synthesizing emotive and expressive speech and, to some extent, selecting a target prosodic rendition, simply by using an appropriate reference. However, whilst PT models are used and evaluated in different-text and different-speaker settings, they are invariably trained in the \textit{same-speaker}, \textit{same-text} setting. This has been shown to lead to \textit{acoustic feature entanglement} \cite{pt_main} and \textit{source-speaker leakage} \cite{karlapati2020copycat} where the models do not fully preserve the target-speaker or target verbal content. Several recent approaches try to address the \textit{feature entanglement} problem \cite{battenberg2019effective, karlapati2020copycat, zhang2019learning} but none address the discrepancy between how these models are trained and how they are used during inference.

In the current work, we train a PT model in both different-speaker and different-text settings by using prosodically similar reference utterances from the training data. This novel training regime has two strong motivations. First, it replicates the inference conditions which should lead to better model generalization. Second, it should encourage the prosody encoder to only model prosodic features and not fine-grained acoustic detail, speaker identity, or the verbal component. Our results indicate that PT models do \textit{not} learn a transferable representation of prosody, but something else.

\section{Related Work} \label{sec:related}
Most prosody transfer approaches follow \cite{pt_main} and normally comprise a core acoustic model, most commonly \textit{Tacotron} \cite{shen2018natural}, and a prosody encoder. The acoustic model and the prosody encoder are jointly trained. The prosody encoder produces a fixed-size latent prosody embedding from the reference mel-spectrogram which is combined with the intermediate acoustic model representation then passed to the decoder which generates the output mel-spectrogram. In \cite{pt_main}, the joint model is trained to minimize target mel-spectrogram loss, which can be viewed as a \textit{reconstruction loss}. In \cite{pt_main} the authors find that this approach is capable of capturing acoustic correlates of prosody from the reference, even if spoken by an unseen speaker or containing verbal content not related to the target-text.

The information capacity of the latent prosody representation has an important effect in the downstream TTS task. Representations that are lower-capacity than that in \cite{pt_main}, such as \cite[for example]{wang2018style, akuzawa2018expressive, zhang2019learning}, are effectively bottle-necked by their capacity and cannot model the same level of detail as \cite{pt_main}. These models are used for tasks such as speaking-style or emotion modelling. Representations of higher capacity than \cite{pt_main}, such as \cite{Lee_Kim_2019, karlapati2020copycat}, capture a considerably higher-detailed representation of the reference and are highly reference-specific. Therefore, high-capacity representations are used for tasks such as voice-puppetry where we expect the lexical content of the reference to be similar to the target text.

Recently, \textit{Daft-Exprt} \cite{zaidi2021daft} reported strong results in a different-text, different-speaker PT task, while maintaining comparatively high perceived naturalness. Daft-Exprt is compared to GST-Tacotron \cite{wang2018style}, VAE-Tacotron \cite{zhang2019learning} and Flowtron \cite{valle2020flowtron} and is rated significantly better than each one in the PT task. All models, including Daft-Exprt, generate a fixed-size representation of prosody similarly to \cite{pt_main}. The information capacity of the representation of prosody is also similar to \cite{pt_main}. 

But Daft-Exprt is different from the other models compared in several ways. Daft-Exprt is based on FastSpeech~2 \cite{ren2020fastspeech} while most other work on PT uses a Tacotron architecture \cite[for example]{pt_main, zhang2019learning, battenberg2019effective, Lee_Kim_2019, klimkov2019fine}. Typically, the prosody embedding is incorporated into the acoustic model using concatenative or additive conditioning \cite[for example]{pt_main, Lee_Kim_2019}. Daft-Exprt instead uses \textit{FiLM} layers \cite{perez2018film} to incorporate the reference prosody representation. Daft-Exprt also uses a jointly-trained speaker classifier with gradient reversal \cite{ganin2015unsupervised} to mitigate source-speaker leakage: the prosody encoder is penalized for learning a latent prosody representation that contains source-speaker information. It is found that when the influence of the speaker classifier -- $\lambda_a$ in Equation \ref{eq:1} -- is increased, the model better preserves target-speaker identity but at the cost of perceived prosody transfer.

There have been several approaches, like Daft-Exprt, recently that try to address source-speaker leakage and feature entanglement in PT by changing the model. Daft-Exprt uses a speaker classifier with gradient reversal to flush out information about the source speaker from the prosody representation. In \cite{battenberg2019effective}, the modelling of the prosody embedding is made conditionally dependent on not only the reference spectrogram but also the reference speaker and text. This is thought to improve performance in different-speaker or different-text PT tasks. Variational auto-encoders (VAE) have also been proposed for reference encoding \cite{zhang2019learning, karlapati2020copycat}. VAEs learn a \textit{disentangled} representation of the input, leading to a tighter control over the effect it has in the downstream TTS task. 

\section{Method}
The approaches mentioned in Section \ref{sec:related} assume that the underlying method generates \textit{transferable representations of prosody}. Using this assumption, we suggest changing how a PT model is trained rather than changing the model to address the frequent problems in PT. Instead of using the target utterance as the reference during training, we employ a reference that is \textit{different} but that should be informative about the target's prosody. We try two simple ways of selecting a suitable reference for each target utterance in the training set.

\subsection{Text-based method}
There is a correlation between the verbal and prosodic components of speech but predicting this similarity is not straightforward. Different renditions of the same text are, however, likely to have a similar prosodic structure. Therefore we consider the case where the reference utterance is of the same text as the target utterance but read by a different speaker than the target-speaker. We use a parallel speech corpus \cite{ribeiro2018parallel} to create such (reference, target) pairs. Of course, one speaker's interpretation of the text will not be identical to another's, leading to prosodic differences. But we assume that the reference will still be \textit{informative} for the PT model and help it predict the target prosody. We refer to the model trained with these data as \textit{\textbased}.

\subsection{\fzbased method}
To compare to the above text-based selection of the reference, we devised a method based on \F0 similarity because \F0 is the principal acoustic correlate of prosody, being influenced by both local and supra-segmental prosodic characteristics. We measure \F0 similarity after aligning utterances using Dynamic Time Warping (DTW), as in \cite{rilliard2011using}, then select the single closest reference for each target utterance in the training set. The model trained with (reference, target) pairs selected in this way is referred to as \textit{\fzbased}.

\subsection{Baseline model}
\begin{figure}[t]
  \centering
  \includegraphics[width=0.9\linewidth]{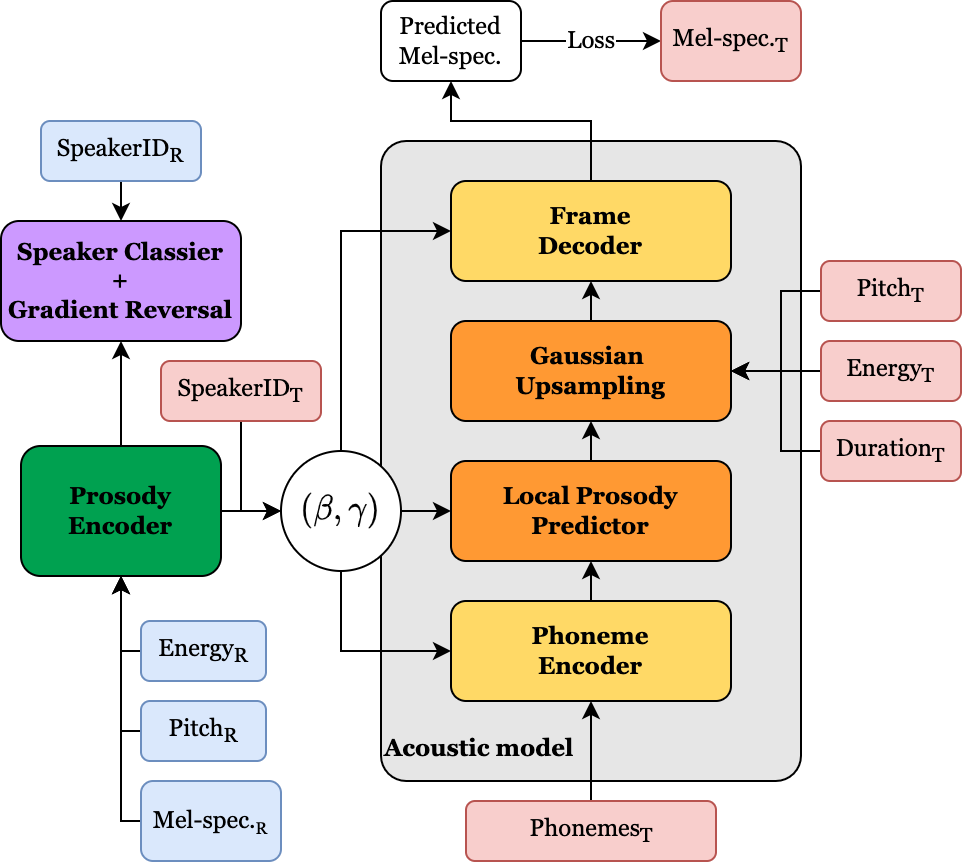}
  \caption{Daft-Exprt training involves a target utterance (information shown in red and indicated with "T") and a reference utterance (information shown in blue, indicated with "R").}
  \label{fig:model_diagram}
\end{figure}

We choose Daft-Exprt as our baseline. Figure \ref{fig:model_diagram} shows how this model employs the reference utterance information during training. The prosody encoder predicts a prosody embedding from the reference mel-spectrogram. The speaker classifier tries to predict the \textit{reference} speaker from that prosody embedding, with gradient reversal discouraging the prosody embedding from containing reference-speaker identity. Separately, the \textit{target} speaker identity is summed to the prosody embedding to indicate which voice to synthesize. 

There are FiLM-conditioning layers at strategic locations in the acoustic model -- phoneme encoder, local prosody predictor, and frame decoder -- and all of their $(\gamma, \beta)$ FiLM parameters are predicted from the prosody embedding by a linear layer. The prosody embedding of the reference thus influences the learned phoneme representation, predicted \F0, energy, duration, and mel-spectrogram prediction. The model is trained to minimize a loss comprising 4 terms:
\begin{align}
    \mathcal{L} = \mathcal{L}_e + \lambda_f\mathcal{L}_f - \lambda_a\mathcal{L}_a + \lambda_r\mathcal{L}_r \label{eq:1}
\end{align}
\noindent where $\mathcal{L}_f$ regularizes the scale of $(\gamma, \beta)$ FiLM parameters as in \cite{oreshkin2018tadam} and $\mathcal{L}_r$ corresponds to weight decay. $\mathcal{L}_a$ is the speaker classifier loss, the $(-)$ indicating gradient reversal. Duration, energy, \F0 and mel-spectrogram losses are combined in $\mathcal{L}_e$, which is the same as the standard loss for the core acoustic model \cite{ren2020fastspeech} with the addition of mean absolute mel-spectrogram error.

\section{Evaluation}
We evaluated naturalness, preservation of target-speaker identity, and prosody transfer for the two training schemes and two additional schemes: \daft trained as in \cite{zaidi2021daft} where the reference and target are identical during training; \shuffle where the reference is a randomly-selected utterance, which is unlikely to be informative about the target utterance prosody.

To create the stimuli to be used in the listening test, we selected 60 test sentences never seen in training. 30 of these were paired with a same-text reference utterance and the other 30 with a different-text reference utterance. No reference utterance was used as either target or reference during training of any model. A random target speaker was assigned to each test sentence. We then synthesized each sentence (using its given reference and assigned target speaker) 4 times, once with each of 4 models: \textbased, \fzbased, \daft, and \shuffle.

\subsection{Model training}
We trained all models on the \textit{Parallel audiobook corpus} \cite{ribeiro2018parallel} which was created from LibriVox and comprises over 120 hours of speech from 59 speakers. All utterances in the corpus are parallel: every sentence\footnote{We use this term for simplicity, although they are not necessarily grammatically-complete sentences, due to the way the data was segmented.} is read by multiple speakers. We removed all sentences longer than 200 characters, leaving 75,267 spoken renditions of 16,275 different sentences. Following \cite{zaidi2021daft}, 80 bin mel-spectrograms were extracted from recordings and phoneme durations were found using MFA \cite{mcauliffe2017montreal}. Log-\F0 was estimated using REAPER\footnote{https://github.com/google/REAPER} and energy defined as the $l^2$-norm of spectrogram frames. Energy and log-\F0 were normalized per speaker. For \fzbased, we used DTW to align sequences of per-phone speaker-normalized log-\F0 to select the closest reference for each utterance in the training corpus. We limited the search to utterances that differ by no more than $\pm$15\% in phone sequence length. After finding the most similar reference for each utterance, we eliminated any pair with a DTW distance greater than 1 standard deviation above the mean DTW distance across the corpus. This resulted in approximately 55,000 utterance pairs.

We trained each model for 24 hours on 16 NVIDIA A100-SXM-80GB GPUs with a batch size of 192. We fine-tuned a pre-trained HiFi-GAN \cite{kong2020hifi} vocoder for each Daft-Exprt model we trained. 

\subsection{Evaluation setup}
For consistency with the PT literature, perceived naturalness of all methods and ground truth utterances was evaluated with a Mean Opinion Score (MOS) test. We evaluated each model's synthesis of 30 of the test sentences, equal amounts of same-text and different-text, resulting in 120 evaluations across the four models. We also evaluated 30 ground truth utterances, both raw and vocoded, using HiFi-GAN to gauge possible reduction in perceived naturalness resulting from vocoding. This results in 180 MOS screens in total.

To evaluate target-speaker identity preservation, we used a discriminative AXY test where listeners indicate whether synthesized sample A sounds more like a ground-truth sample from target-speaker X or a ground-truth sample from the reference speaker Y. We used the different-speaker samples from our test set, resulting in 30 speaker AXY screens per model $=$ 120 in total.

Various objective and subjective metrics are used in \cite{pt_main} to evaluate prosody transfer. These include metrics such as mel cepstral distortion \cite{kubichek1993mel} and voicing decision error \cite{nakatani2008method}. Such objective metrics require a gold-standard against which to calculate distortion or error, which is only possible for same-text, same-speaker PT. Since that is not possible here, we followed \cite{zaidi2021daft} and used a MUSHRA-like test for evaluating PT, using \shuffle as the anchor. Like in \cite{pt_main}, we asked listeners to focus on how pitch changes, word stress, speaking rate, and pause lengths. We evaluated all same-text and different-text samples from our test set, resulting in 60 MUSHRA screens, each including one sample from each of the 4 trained models.

Each screen was rated by at least 8 different English-speaking listeners based in either the US or the UK recruited via Prolific\footnote{https://www.prolific.co/}. Each participating listener completed 36 screens, pseudo-randomly chosen from each evaluation category.

\section{Results}

\subsection{Naturalness} \label{sec:naturalness}
MOS scores in Table \ref{tab:mos-results} show a significant decrease, according to a paired t-test, in perceived naturalness when audio from the corpus is vocoded using a HiFi-GAN fine-tuned on ground-truth audio.  \textbased and \fzbased are statistically not different in naturalness from \shuffle for same-text PT, but significantly worse for different-text PT. Neither \fzbased nor \textbased are rated significantly better in same-text PT than in different-text PT. \daft is rated significantly better than \shuffle only for same-text PT, and significantly \textit{worse} than \shuffle for different-text PT.

\begin{table}[th]
\caption{MOS results for both real and synthesized samples. Target-speakers are chosen at random.} \label{tab:mos-results}
\centering
\begin{tabularx}{\columnwidth}{lll}
\toprule
\multicolumn{1}{c}{\textbf{Model}} & \multicolumn{2}{c}{\textbf{MOS}}.      \\
\midrule
Ground Truth               & \multicolumn{2}{c}{$4.2\pm0.1$}               \\
Ground Truth + HiFi-GAN    & \multicolumn{2}{c}{$3.7\pm0.1$}               \\
\midrule
                           & \multicolumn{1}{c}{\textbf{Same-text}}             & \multicolumn{1}{c}{\textbf{Different-text}}             \\
\midrule
\shuffle                & $2.8\pm0.2$          & $2.9\pm0.2$          \\
\textbased          & $2.6\pm0.2$          & $2.4\pm0.2$          \\
\fzbased                & $2.9\pm0.2$          & $2.6\pm0.2$          \\
\midrule
\daft             & $3.2\pm0.2$          & $2.4\pm0.2$          \\
\bottomrule
\end{tabularx}
\end{table}

\subsection{Target speaker identity} \label{sec:speaker}
Table \ref{tab:mushra-intra-results} shows listeners' AXY classification accuracy for the four evaluated models. Excluding \daft, all models have a high speaker classification accuracy, indicating that these models successfully preserve the target speaker identity in different-speaker PT. \daft on the other hand has a very low target speaker classification accuracy. In fact, the reference speaker is chosen more frequently than the target speaker by listeners: there is substantial source-speaker leakage in \daft.

\subsection{Prosody transfer}
MUSHRA-like scores are reported for same-text and different-text PT in Table \ref{tab:mushra-intra-results} with their 95\% confidence intervals. \daft is significantly better than all models under each condition. \daft is, not surprisingly, significantly better under same-text conditions than different-text conditions as this matches its training condition.

\begin{table}[th]
\caption{PT MUSHRA-like scores and target-speaker classification accuracy. Target-speakers are randomly sampled.} \label{tab:mushra-intra-results}
\centering
\begin{tabularx}{\columnwidth}{llll}
\toprule
\multicolumn{1}{c}{\multirow{2}{*}{\textbf{Model}}} & \multicolumn{2}{c}{\textbf{MUSHRA-like}} & \multicolumn{1}{c}{\textbf{Speaker}} \\
\multicolumn{1}{c}{} & \multicolumn{1}{c}{\textbf{Same-text}}  & \multicolumn{1}{c}{\textbf{Different-text}} & \multicolumn{1}{c}{\textbf{classif.}}\\
\midrule
\shuffle               & $39.0\pm4.2$  & $25.5\pm3.1$  & \multicolumn{1}{r}{$91.5\%$}\\
\textbased             & $38.7\pm4.4$  & $30.4\pm3.2$  & \multicolumn{1}{r}{$88.4\%$}\\
\fzbased               & $42.9\pm4.5$  & $28.4\pm3.1$  & \multicolumn{1}{r}{$91.0\%$}\\
\midrule
\daft.                 & $61.5\pm4.8$  & $49.3\pm4.1$  & \multicolumn{1}{r}{$46.1\%$}\\
\bottomrule
\end{tabularx}
\end{table}

\subsection{Analysis}
Our results indicate that the two proposed training methods, \textbased and \fzbased, have a negative impact on PT performance. To better understand why participants prefer \daft we objectively analyzed the predicted \F0 contours for all models with two metrics: frame-level DTW \F0 alignment and mean absolute \F0 error to evaluate how well they preserve the mean \F0 of the target-speaker. The results are shown in Table \ref{tab:objetive-metrics}, averaged over the 60 sentence test set. The \F0 contour predicted by \daft is most similar to the reference in all PT conditions, indicating that the other methods do not capture the reference \F0 contour shape as well as \daft. 

In same-speaker PT, all models accurately capture the mean \F0 of the target speaker. In the different-speaker case, however, \daft performs significantly worse than other models. We find that the mean \F0 of the output is always driven towards the mean \F0 of the reference as evidenced in our results in Section \ref{sec:speaker}.

\begin{table}[th]
\caption{A normalized DTW-based metric indicates how well \F0 contours align with the reference \F0 contour (lower is better, $0$ indicating perfect alignment, $1$ indicates worst alignment). Mean absolute F0 error indicates how well each model preserves the target-speaker identity.} \label{tab:objetive-metrics}

\centering
\begin{tabular}{lllll}
\toprule
           & \multicolumn{2}{c}{\textbf{\F0 DTW error}} & \multicolumn{2}{c}{\textbf{Mean \F0 target error}} \\
           & same spkr.      & diff spkr.     & same spkr.             & diff spkr.             \\
\midrule
\shuffle   & 0.60             & 0.90           & \SI{19.9}{Hz}          & \SI{20.6}{Hz}          \\
\textbased & 0.60             & 0.95           & \SI{16.9}{Hz}          & \SI{18.2}{Hz}          \\
\fzbased   & 0.50             & 0.85           & \SI{25.7}{Hz}          & \SI{20.9}{Hz}          \\
\midrule
\daft      & 0.35             & 0.45           & \SI{25.4}{Hz}          & \SI{43.5}{Hz}          \\
\bottomrule
\end{tabular}
\end{table}
We found that a PT model trained using the proposed methods performs similarly to \shuffle. A possible explanation for these results is that our methods for choosing references during training do not actually select prosodically similar utterances. So, we performed an additional analysis to test for this. We asked 55 participants to indicate which of three references (selected using \textbased, \fzbased and \shuffle), is most prosodically similar to a target utterance. Participants were asked to focus on the same prosodic aspects as in our previous evaluation of PT. Each participant performed 16 ratings resulting in 880 responses. Different from our PT evaluation, this survey asked participants to compare stimuli with different lexical content. To focus participants on only the prosodic content of the stimuli we delexified all samples using an attenuated low-pass filter with a cut-off at $\SI{200}{\hertz}$ and a roll-off of $\SI{24}{\decibel}$ per octave.

\begin{figure}[t]
  \centering
  \includegraphics[width=0.9\linewidth]{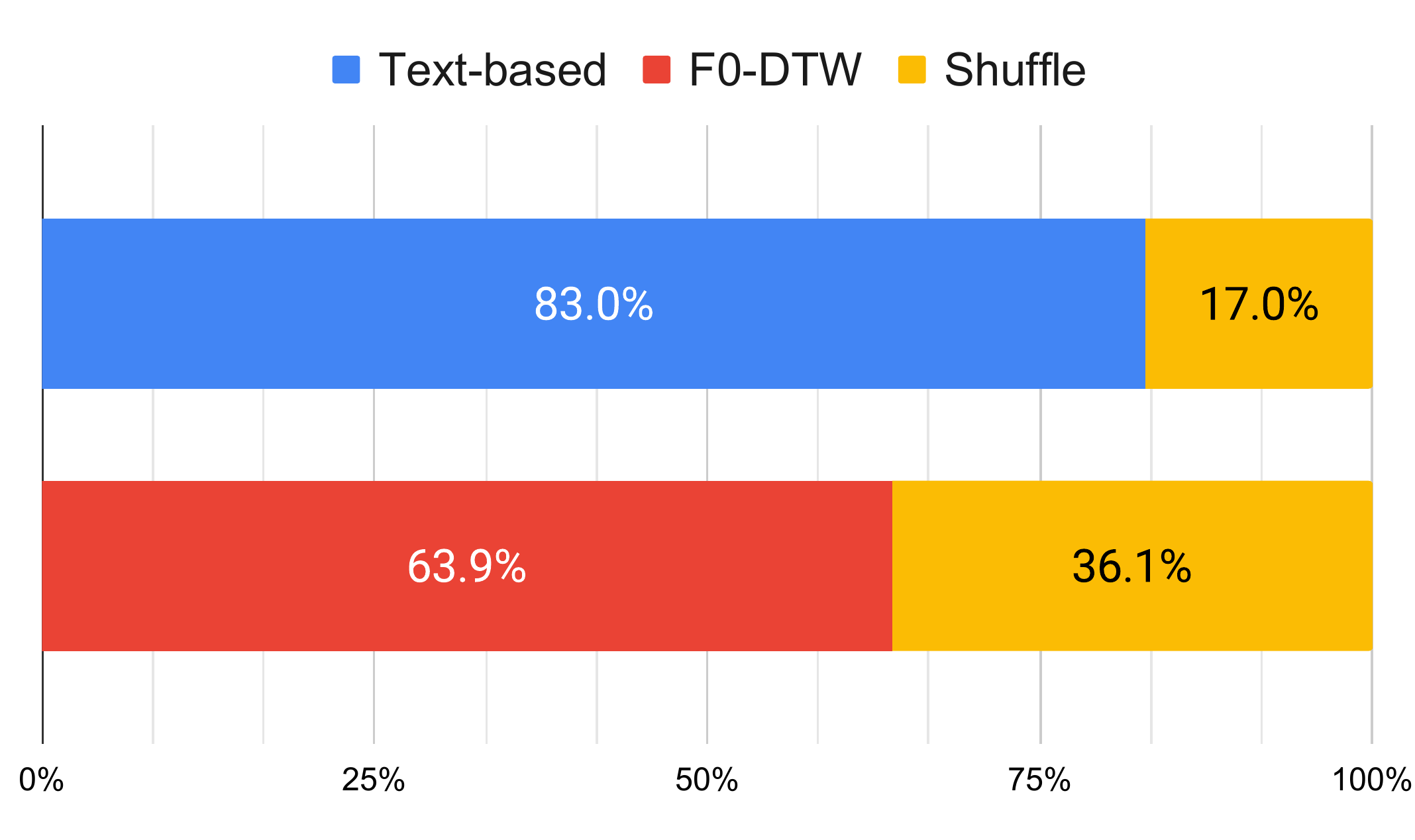}
  \caption{Our evaluation task shows that evaluators prefer the two proposed methods over randomly sampled utterances.}
  \label{fig:preference}
\end{figure}

The results of this analysis, shown in Figure \ref{fig:preference}, indicate a clear preference of both proposed methods when compared to randomly sampled utterances. In other words, our proposed methods do indeed select prosodically similar references for training the model. The low PT quality of \textbased and \fzbased can therefore not be explained by low prosodic similarity in the training utterance pairs.

\section{Conclusions}

If prosody transfer models did actually transfer prosody it should be possible to train them using prosodically similar references, rather than with a reference \textit{identical} to the target. Our results indicate that this is not the case. As can be seen in the results of Sections \ref{sec:naturalness}-\ref{sec:speaker}, \daft suffers from source-speaker leakage in the different-speaker condition, and significantly lower perceived naturalness in the different-text condition.

Based on our results we conclude that \daft stands out to listeners in the PT task because it is the only model that transfers prosodically-important acoustic features, like \F0, to make them highly similar to the reference. This similarity to the reference indicates that \daft models a representation of prosody that is highly dependent on the reference text, regardless of the target text. However, a model trained on references with high \F0 alignment (\fzbased) is rated similar to \shuffle. Furthermore, \daft is rated highest in same-text, different-speaker PT while a model trained under these conditions (\textbased) performs similarly to an uninformed model (\shuffle). This suggests that the prosodic representation modeled by \daft is highly reference-speaker dependent.

Prosody is text- and speaker-dependent. Therefore, a transferable representation of prosody has to be invariant to the reference speaker and reference text so that it can be applied to any target text and speaker. We confirmed that utterance pairs selected by \textbased and \fzbased are prosodically similar  but this similarity does not result in good prosody transfer performance.  Models such as \daft therefore require that the reference is identical to the target during training. Since the representation modeled by \daft is dependent on both the reference speaker and reference text we conclude that the prosodic representation modeled by \daft is not transferable.

\bibliographystyle{IEEEbib}
\bibliography{mybib}

\end{document}